\documentclass{article}
\usepackage{spconf,amsmath,graphicx}
\usepackage{amsfonts}
\usepackage{algorithmic}
\usepackage{algorithm}
\usepackage{array}
\usepackage[caption=false,font=normalsize,labelfont=sf,textfont=sf]{subfig}
\usepackage{textcomp}
\usepackage{stfloats}
\usepackage{url}
\usepackage{verbatim}
\usepackage{cite}
\usepackage{acronym, color, soul}
\usepackage{hyperref}
\usepackage{booktabs}
\usepackage{multirow, bm}
\usepackage{adjustbox}
\usepackage{xspace}

\makeatletter
\DeclareRobustCommand\onedot{\futurelet\@let@token\@onedot}
\def\@onedot{\ifx\@let@token.\else.\null\fi\xspace}

\def\etal{\emph{et al}\onedot}

\newacro{rnn}[RNN]{recurrent neural network}
\newacro{ae}[AE]{autoencoder}
\newacro{resblock}[ResBlock]{residual block}

\newacro{vvc}[VVC]{versatile video coding}
\newacro{hevc}[HEVC]{high-efficiency video coding}
\newacro{bpg}[BPG]{better portable graphics}

\newacro{charm}[ChARM]{channel-wise auto-regressive model}
\newacro{rpn}[ResizeParamNet]{resize parameter network}
\newacro{stn}[STN]{spatial transformer network}
\newacro{tfc}[TFC]{tensorflow compression}
\newacro{aict}[AICT]{adaptive image compression transformer}
\newacro{ict}[ICT]{image compression transformer}
\newacro{nic}[NIC]{neural image compression}
\newacro{swint}[SwinT]{swin transformer}
\newacro{rd}[RD]{rate-distortion}
\newacro{flops}[FLOPs]{floating point operations per second}

\title{AICT: An Adaptive Image Compression Transformer}

\name{Ahmed Ghorbel$^{1}$ \qquad  Wassim Hamidouche$^{1,2}$ \qquad Luce Morin$^{1}$ \thanks{This work has been supported by Région Bretagne and Rennes Ville et Métropole under the DEEPTEC project.}}
\address{$^1$ Univ Rennes, INSA Rennes, CNRS, IETR – UMR 6164, F-35000 Rennes, France \\
      $^{2}$ Technology Innovation Institute P.O.Box: 9639, Masdar City Abu Dhabi, UAE}

\begin{document}
\ninept
\maketitle

\begin{abstract}
Motivated by the efficiency investigation of the Tranformer-based transform coding framework, namely SwinT-ChARM, we propose to enhance the latter, as first, with a more straightforward yet effective Tranformer-based channel-wise auto-regressive prior model, resulting in an absolute \ac{ict}.
Current methods that still rely on ConvNet-based entropy coding are limited in long-range modeling dependencies due to their local connectivity and an increasing number of architectural biases and priors. On the contrary, the proposed \ac{ict} can capture both global and local contexts from the latent representations and better parameterize the distribution of the quantized latents.
Further, we leverage a learnable scaling module with a sandwich ConvNeXt-based pre/post-processor to accurately extract more compact latent representation while reconstructing higher-quality images.
Extensive experimental results on benchmark datasets showed that the proposed \ac{aict} framework significantly improves the trade-off between coding efficiency and decoder complexity over the \ac{vvc} reference encoder (VTM-18.0) and the neural codec SwinT-ChARM.
\end{abstract}


\acresetall
\vspace{-1mm}
\section{Introduction}
\label{intro}
Visual information is crucial in human development, communication, and engagement, and its compression is necessary for effective data storage and transmission. Thus, designing new lossy image compression algorithms is a goldmine for scientific research. The goal is to reduce an image file size by permanently removing less critical information, specifically redundant data and high frequencies, to obtain the most compact bit-stream representation while preserving a certain level of visual fidelity. Nevertheless, the high compression rate and low distortion are fundamentally opposed objectives involving optimizing the \ac{rd} cost.

Conventional compression standards, including JPEG, JPEG2000, H.265/HEVC, and H.266/VVC, rely on hand-crafted creativity to present module-based encoder/decoder block diagram, i.e., Intra prediction, transform, quantization, arithmetic coding, and post-processing. Traditional coding algorithms have a lot of advantages, including mature technology with SW/HW-friendly implementations, low decoding complexity, and strong generalization on different contents.
\begin{figure}[htb]
\vspace{-1mm}
\centering
\includegraphics[width=0.49\textwidth]{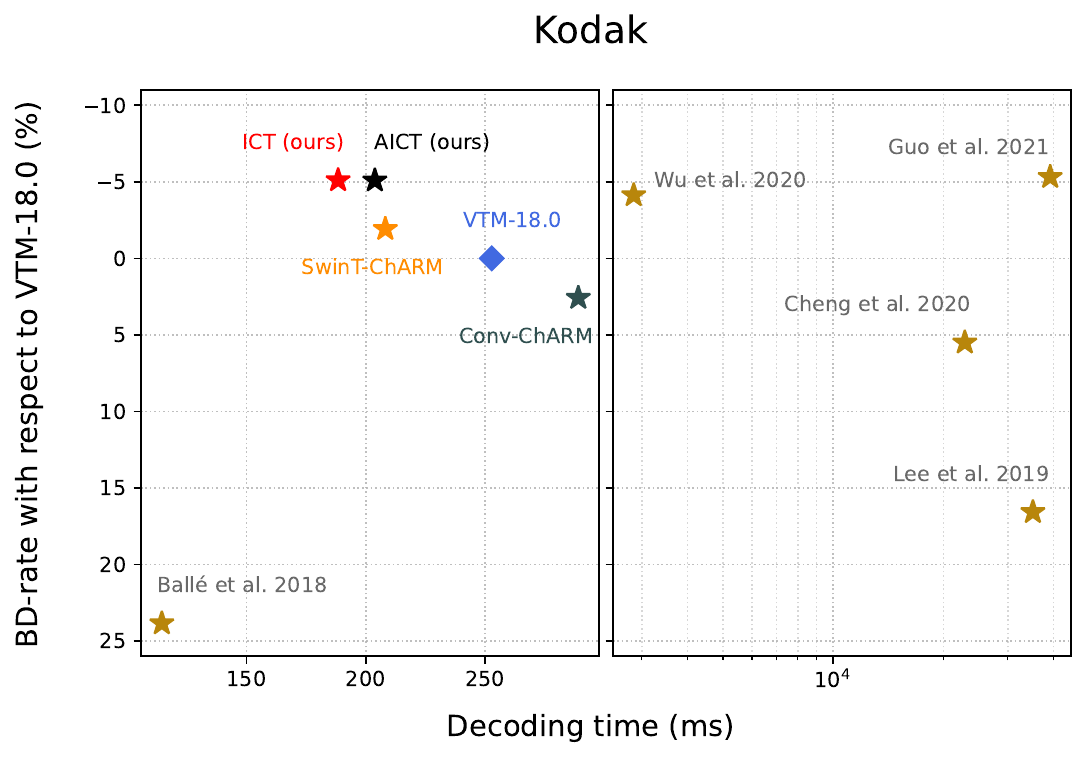}
\vspace{-5mm}
\caption{BD-rate (\%)$\downarrow$ versus decoding time (ms)$\downarrow$ on the Kodak dataset. Left-top is better. Star and diamond markers refer to decoding on GPU and CPU, respectively.}
\vspace{-3mm}
\label{bdr_vs_dt}
\end{figure}
Nevertheless, all of them mainly rely on hand-crafted coding techniques; thus, it is quite challenging to directly optimize \ac{rd} cost for all types of image content due to the rapid development of new image formats and the growth of high-resolution mobile devices.
On the other hand, neural coding has gained wide attention from research and industry, yielding promising end-to-end \ac{nic} solutions outperforming their conventional counterparts in coding efficiency. \Ac{nic} leverages \acp{ae} to carry out a non-linear coding from the signal domain to a compact representation. Such \ac{ae}-based system consists of three modular parts: transform, quantization, and entropy coding, which can be trained in an end-to-end fashion to minimize the distortion between a source image and its reconstruction, and the rate needed to convey the latent representation bit-stream.

Since the early \ac{rnn}-based method \cite{toderici2015variable} for lossy image compression, significant advancements have been made in integrating tailored modules for \ac{nic}. Previous works use local context \cite{minnen2018joint,lee2019extended,mentzer2018conditional}, or additional side information \cite{balle2018variational,hu2020coarse,minnen2018image} to capture short-range spatial dependencies, and others use non-local mechanisms \cite{cheng2020learned,li2021learning,qian2020learning,chen2021end} to model long-range spatial dependencies. Recently, Toderici \etal \cite{toderici2022high} proposed a generative compression method achieving high-quality reconstructions, Minnen \etal \cite{minnen2020channel} introduced channel-conditioning taking advantage of an entropy-constrained model that uses both forward and backward adaptations, Zhu \etal \cite{zhu2021transformer} replaced the ConvNet-based transform coding in the Minnen \etal \cite{minnen2020channel} approach with a Transformer-based one, Zou \etal  \cite{zou2022devil} combined the local-aware attention mechanism with the global-related feature learning and proposed a window-based attention module, Koyuncu et al. \cite{koyuncu2022contextformer} proposed a Transformer-based context model, which generalizes the standard attention mechanism to spatio-channel attention, Zhu \etal \cite{zhu2022unified} proposed a probabilistic vector quantization with cascaded estimation under a multi-codebooks structure, Kim \etal \cite{kim2022joint} exploited the joint global and local hyperpriors information in a content-dependent manner using an attention mechanism, and He \etal \cite{he2022elic} adopted stacked residual blocks as nonlinear transform and multi-dimension entropy estimation model.

In order to improve image-level prediction while minimizing computation costs, learned sampling techniques have been developed for several vision tasks. \Acp{stn}~\cite{jaderberg2015spatial} introduce a layer that estimates a parametrized affine, projective, and splines transformation from an input image to recover data distortions. Based on the latter, Chen \etal \cite{chen2022estimating} proposed a straightforward learned downsampling module that can be jointly optimized with any neural compression kernels in an end-to-end fashion. Talebi \etal \cite{talebi2021learning} jointly optimize pixel value interpolated at each fixed downsampling location for classification. Jin \etal \cite{jin2021learning} introduced a deformation module and a learnable downsampling operation, which can be optimized together with the given segmentation model.

One of the main challenges of \ac{nic} is the ability to identify the crucial information necessary for the reconstruction, knowing that information overlooked during encoding is usually lost and unrecoverable for decoding. Another main challenge is the trade-off between coding performance and decoding latency. While the existing approaches improve the transform and entropy coding accuracy, they still need to be improved by the higher decoding runtime and excessive model complexity leading to an ineffective real-world use. To cope with those challenges, we present in this paper three contributions summarized as follows:
\vspace{-1mm}
\begin{itemize}
\item We propose the \ac{ict}, a nonlinear transform coding and spatio-channel auto-regressive entropy coding. These modules are based on \ac{swint} blocks for effective latent decorrelation and a more flexible receptive field to adapt for contexts requiring short/long-range information.
\item We further propose the \ac{aict} model that adopts a scale adaptation module as a sandwich processor to enhance compression efficiency. This module consists of a neural scaling network and ConvNeXt-based pre/post-processor to jointly optimize differentiable resizing layers and a content-dependent resize factor estimator.
\item We conduct experiments on four widely-used benchmark datasets to explore possible coding gain sources and demonstrate the effectiveness of \ac{aict}. In addition, we carried out a model scaling analysis and an ablation study to substantiate our architectural decisions.
\end{itemize}
%
Extensive experiments reveal the impacts of the spatio-channel entropy coding, the sandwich scale adaptation component, and the joint global structure and local texture learned by the self-attention units through the nonlinear transform coding. These experiments validate that the proposed \ac{aict} model achieves compelling compression performance, as illustrated in Fig.~\ref{bdr_vs_dt}, outperforming conventional and neural codecs in both coding efficiency and decoder complexity. 

The rest of this paper is organized as follows. First, the proposed \ac{aict} framework is described in detail in Section~\ref{methd}. Next, we dedicate Section~\ref{result} to describe and analyze the experimental results. Finally, Section~\ref{conclusion} concludes the paper.

%
\begin{figure*}[htb]
\centering
\includegraphics[width=1\textwidth]{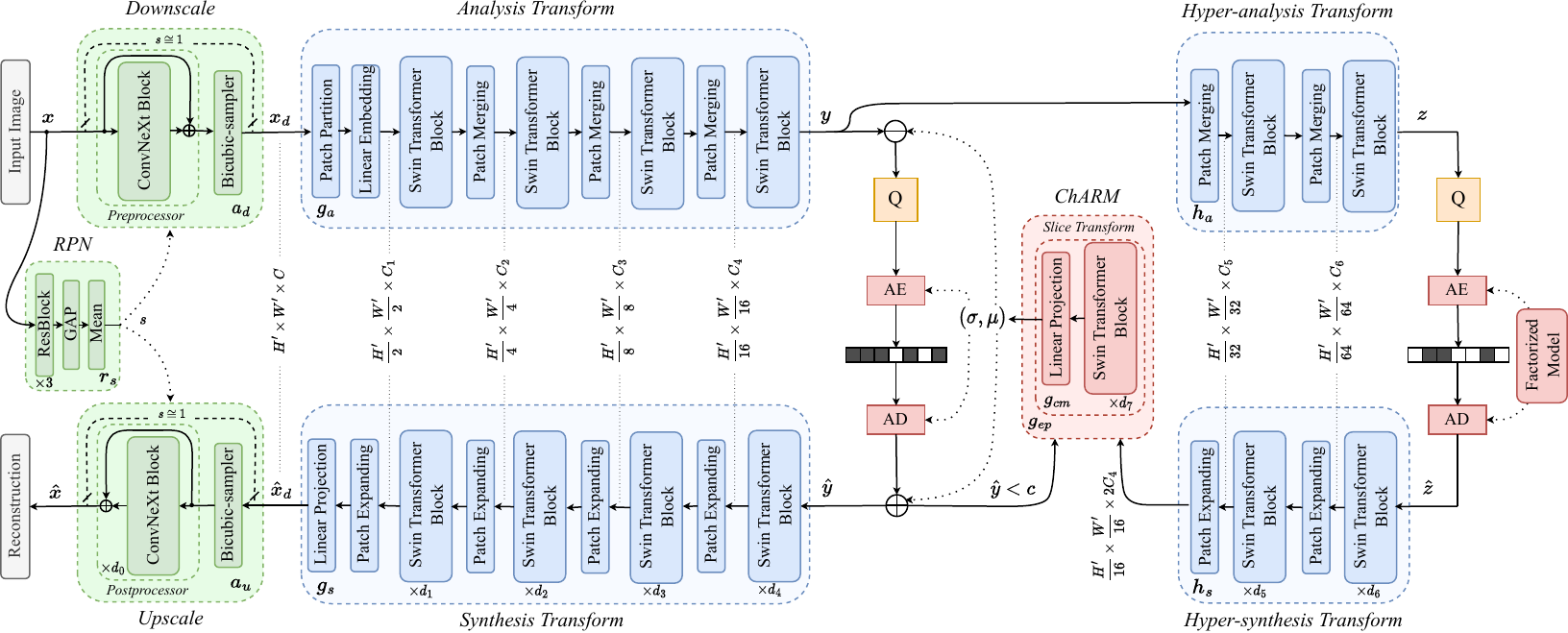}
\caption{Overall \acs{aict} framework. We illustrate the image compression diagram of our \acs{aict} with hyperprior and SwinT-based \ac{charm}, and scale adaptation module. The \ac{rpn} and ConvNeXt block diagrams are illustrated in Fig.~\ref{blocks} (a) and (c).}
\label{ov_framework}
\end{figure*}
\section{Proposed AICT Framework}
\label{methd}
\subsection{Overall Architecture} 
The overall pipeline of the proposed solution is illustrated in Fig.~\ref{ov_framework}. The framework includes three modular parts. First, the scale adaptation module, composed of a tiny \ac{rpn} \cite{chen2022estimating}, a ConvNeXt-based pre/post-processor, and a bicubic interpolation filter. Second, the analysis/synthesis transform $(g_{a},g_{s})$ of our design consists of a combination of patch merging/expanding layers and \ac{swint}~\cite{liu2021swin} blocks. The architectures of hyper-transforms $(h_{a},h_{s})$ are similar to $(g_{a},g_{s})$ with different stages and configurations. Finally, a Transformer-based slice transform under a \ac{charm} design is used to estimate the distribution parameters of the quantized latent. These resulting discrete-valued data $(\hat{y}, \hat{z})$ are encoded into bit-streams with an arithmetic coder.

\vspace{-1mm}
\subsection{Scale Adaptation Module}
\begin{figure}[htbp!]
\centering
\includegraphics[width=0.48\textwidth]{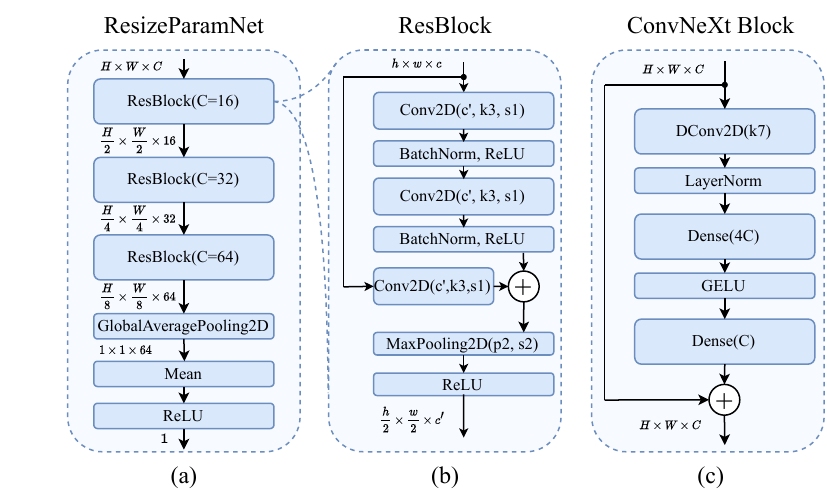}
\caption{Adaptation module block architectures.}
\label{blocks}
\end{figure}

Given a source image $x \in R^{H \times W \times C}$, we first determine an adaptive resize factor $M$ estimated by the \ac{rpn} module, which consists of three stages of \acp{resblock}. Indeed, the estimated resize parameter $M$ is used to create a sampling grid $\tau_{M}$ following the convention \acp{stn}, and used to adaptively down-scale $x$ into $x_{d} \in R^{H' \times W' \times C}$ through the bicubic interpolation. The latter is then encoded and decoded with the proposed \ac{ict}. Finally, the decoded image $\hat{x}_{d} \in R^{H' \times W' \times C}$ is up-scaled to the original resolution $\hat{x} \in R^{H \times W \times C}$ using the same, initially estimated, resize parameter $M$.
The parameterization of each layer is detailed in the \ac{rpn} and \ac{resblock} diagrams of Fig.~\ref{blocks} (a) and (b), respectively. In addition, a learnable depth-wise pre/post-processor is placed before/after the bicubic sampler to mitigate the information loss introduced by down/up-scaling, allowing the retention of information. This neural pre/post-processing method consists of concatenation between the input and the output of three successive ConvNeXt~\cite{liu2022convnet} blocks. The ConvNeXt block diagram is also illustrated in Fig.~\ref{blocks} (c). For a better complexity-efficient design, we decided to skip the scale adaptation module where $M \cong 1$.

\vspace{-1mm}
\subsection{Transformer-based Analysis/Synthesis Transform}
The analysis transform $g_a$ contains four stages of patch merging layer and \ac{swint} block to obtain a more compact low-dimensional latent representation $y$. In order to consciously and subtly balance the importance of feature compression through the end-to-end learning framework, we used two additional stages of patch merging layer and \ac{swint} block in the hyper-analysis transform to produce an additional latent representation $z$. 
During training, both latents $y$ and $z$ are quantized using a rounding function to produce $\hat{y}$ and $\hat{z}$, respectively.
The quantized latent variables $\hat{y}$ and $\hat{z}$ are then entropy coded regarding an indexed entropy model for a location-scale family of random variables parameterized by the output of the \ac{charm}, and a batched entropy model for continuous random variables, respectively, to obtain the bit-streams. Finally, quantized latents $\hat{y}$ and $\hat{z}$ feed the synthesis and hyper-synthesis transforms, respectively, to generate the reconstructed image. The decoder schemes are symmetric to those of the encoder, with patch-merging layers replaced by patch-expanding layers.

\vspace{-1mm}
\subsection{Transformer-based Slice Transform}
Although there are strong correlations among different channels in latent space, the strongest correlations may come from the spatio-channel dependencies.
Thus, to better parameterize the distribution of the quantized latents with a more accurate and flexible entropy model and without increasing the compression rate, we propose a Transformer-based slice transform inside the \ac{charm}. Unlike previous works, ours considers spatio-channel latent correlations for entropy modeling in an auto-regressive manner. As a side effect, it also leads to faster decoding speed.
The slice transform consists of two successive \ac{swint} blocks with an additional learnable linear projection layer, used to get a representative latent slices concatenation. This \ac{charm} estimates the distribution $p_{\hat{y}} (\hat{y} | \hat{z})$ with both the mean and standard deviation of each latent slice, and incorporates an auto-regressive context model to condition the already-decoded latent slices and further reduce the spatial redundancy between adjacent pixels.

\vspace{-5mm}
\section{Results and Analysis}
\label{result}
\subsection{Experimental Setup}
\label{experiments}
{\bf Baselines.\footnote{For a fair comparison, we only considered SwinT-ChARM \cite{zhu2021transformer} from the state-of-the-art models \cite{zhu2021transformer,koyuncu2022contextformer,zou2022devil,zhu2022unified,kim2022joint,he2022elic}, due to the technical feasibility of models training and evaluation under the same conditions and in an adequate time.}}
We compare our solution with the state-of-art neural codec SwinT-ChARM proposed by Zhu \etal \cite{zhu2021transformer}, and the Conv-ChARM proposed by Minnen \etal \cite{minnen2020channel} and conventional codecs, including \ac{bpg}(4:4:4), and the \ac{vvc} official Test Model VTM-18.0 in All-Intra configuration.
\\\\
{\bf Implementation details.}
We implemented all models in TensorFlow using \ac{tfc} library, and the experimental study was carried out on an RTX 5000 Ti GPU and an Intel(R) Xeon(R) W-2145 @ 3.70GHz CPU. All models were trained on the same CLIC20 training set with 2M iterations using the ADAM optimizer with parameters $\beta_1=0.9$ and $\beta_2=0.999$. The initial learning rate is set to $10^{-4}$ and drops to $10^{-5}$ for the last 200k iterations, and $L=D+\lambda{R}$ is used as a loss function. $L$ is a weighted combination of bitrate $R$ and distortion $D$, with $\lambda$ being the Lagrangian multiplier steering \ac{rd} trade-off. Mean squared error (MSE) is used as the distortion metric in RGB color space. Each training batch contains eight random crops $\in R^{256 \times 256 \times 3}$ from the CLIC20 training set. To cover a wide range of rate and distortion points, for our proposed method and respective ablation models, we trained four models with $\lambda \in \{1000, 200, 20, 3\} \times 10^{-5}$. We evaluate the image codecs on four datasets \cite{testsets}, including Kodak, Tecnick, JPEG-AI, and the testing set of CLIC21. For a fair comparison, all images are cropped to the highest possible multiples of 256 to avoid padding for neural codecs.
\begin{figure*}[htbp!]
  \centering
  \subfloat[Rate distortion comparison.]{
    \includegraphics[width=0.44\linewidth]{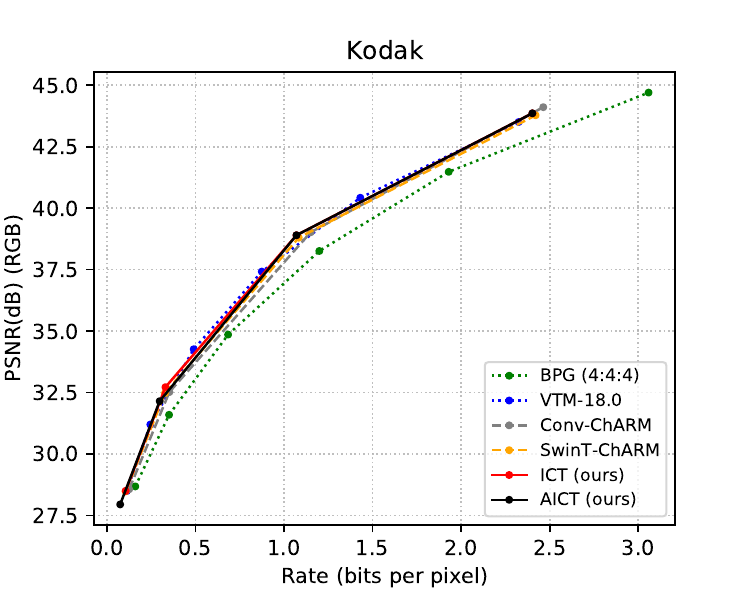}
    \label{rd_curve_kodak}}
  \hfil
  \subfloat[Rate saving over VTM-18.0.]{
    \includegraphics[width=0.42\linewidth]{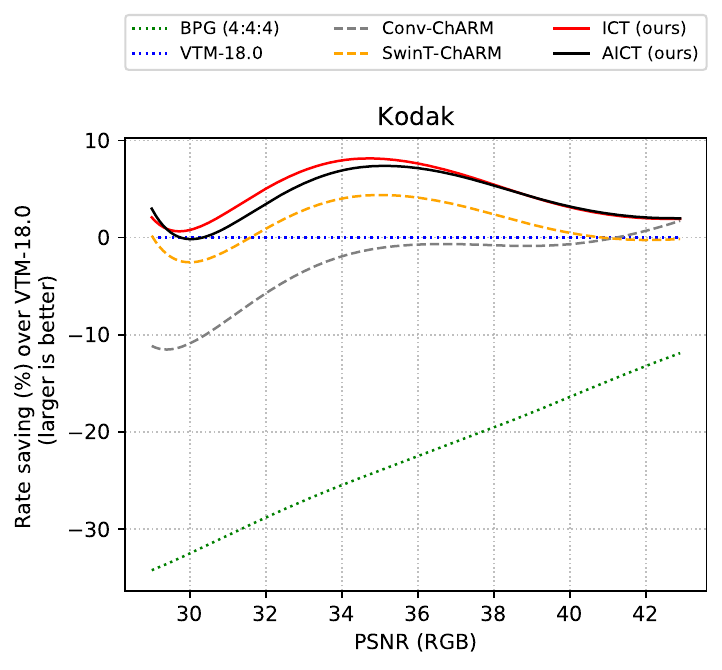}
    \label{rate_savings_kodak}}
  \caption{Compression efficiency comparison on the Kodak dataset.}
  \vspace{-2mm}
  \label{efficiency_kodak}
\end{figure*}

\vspace{-1mm}
\subsection{Rate-Distortion Coding Performance}
To demonstrate the compression efficiency of our proposed approach, we summarize, in Table~\ref{bdrate}, the BD-rate of our models and the baselines across four datasets compared to the VTM-18.0 as the anchor.
On average, \ac{aict} is able to achieve 5.11\% BD-rate reduction compared to VTM-18.0 and 3.93\% relative gain from SwinT-ChARM. Also, we illustrate in Figure~\ref{efficiency_kodak} a comparison of compression efficiency on Kodak dataset.
Figure~\ref{bdr_vs_dt} shows the BD-rate (with VTM-18.0 as an anchor) versus the decoding time of various approaches on the Kodak dataset. It can be seen from the figure that our \ac{ict} and \ac{aict} achieve a good trade-off between BD-rate performance and decoding time.
\begin{table}[!htb]
\centering
\caption{BD-rate$\downarrow$ performance of \ac{bpg} (4:4:4), Conv-ChARM, SwinT-ChARM, \ac{ict}, and \ac{aict} compared to the VTM-18.0.}\label{bdrate}
\adjustbox{max width=0.48\textwidth}{
\begin{tabular}{@{}l|ccccc@{}}
\toprule
Image Codec  & Kodak & Tecnick & JPEG-AI & CLIC21 & {\bf Average }\\
\midrule
BPG444       & 22.28\%          & 28.02\%	       & 28.37\%          & 28.02\% &      26.67\%   \\
Conv-ChARM   &  2.58\%          &  3.72\%          &  9.66\%          &  2.14\%   &    4.53\%  \\
SwinT-ChARM  & -1.92\%          & -2.50\%          &  2.91\%          & -3.22\%    &  -1.18\%   \\
ICT (ours)   & \textbf{-5.10\%} & -5.91\%	       & -1.14\%          & -6.44\%     &  -4.65\%  \\
AICT (ours)  & 
\textbf{-5.09\%} & \textbf{-5.99\%} & \textbf{-2.03\%} & \textbf{-7.33\%} & \textbf{-5.11\%} \\
\bottomrule
\end{tabular}%
}
\end{table}

\vspace{-1mm}
\subsection{Models Scaling Study}
\label{modelscaling}
We evaluated the decoding complexity of the four considered image codecs by averaging decoding time across 7000 images encoded at 0.8 bpp. We present the image codecs complexity in Table~\ref{complexity}, including decoding time on GPU and CPU, codec \ac{flops}, and codec total number of parameters.
Compared to the neural baselines, \ac{ict} can achieve faster decoding speed on GPU but not on CPU, which proves the parallel processing ability to speed up compression on GPU and the well-engineered designs of both transform and entropy coding, highlighting an efficient and hardware-friendly compression model. This is potentially helpful for conducting high-quality real-time visual data streaming. 
Our \ac{aict} is on par with \ac{ict} in terms of the number of parameters, \ac{flops}, and latency, indicating that the scale adaptation module is not computationally heavy for real scenario applications. 
\begin{table}[t]
\centering
\caption{Average decoding latency across 7000 images at 256$\times$256 resolution, encoded at 0.8 bpp. 
}\label{complexity}
\adjustbox{max width=0.48\textwidth}{%
\begin{tabular}{@{}l|cc|c|c@{}}
\toprule
\multirow{2}{*}{Image Codec} & \multicolumn{2}{c|}{{Latency(ms)$\downarrow$}} & \multirow{2}{*}{M\acs{flops}$\downarrow$} & \multirow{2}{*}{\#parameters (M)$\downarrow$}   \\ \cmidrule{2-3}
& GPU  & CPU & & \\
\midrule
Conv-ChARM   & 133.8         & \textbf{359.8} & 126.1999 & 53.8769 \\
SwinT-ChARM  & 91.8          & 430.7          & 63.2143  & 31.3299 \\
ICT (ours)   & \textbf{80.1} & 477.0          & 74.7941  & 37.1324 \\
AICT (ours)  & 88.3          & 493.3          & 74.9485  & 37.2304 \\
\bottomrule
\end{tabular}%
}
\end{table}

\vspace{-1mm}
\subsection{Ablation Study}
\label{ablation}
To investigate the impact of the proposed \ac{ict} and \ac{aict}, we conduct an ablation analysis according to the reported BD-rate results in Table~\ref{bdrate}.
The compression performance increases from Conv-ChARM to SwinT-ChARM on the considered datasets due to the inter-layer feature propagation across non-overlapping windows (local information) and self-attention mechanism (local information) in the \ac{swint}.
With the proposed spatio-channel entropy model, \ac{ict} is able to achieve, on average, -3.47\% BD-rate reduction compared to SwinT-ChARM. Therefore, introducing the Transformer-based slice transform leads to significant improvement compared to the ConvNet-based entropy model using only short-range dependencies. In addition, our spatio-channel entropy model is more helpful when combined with the Transformer-based transform coding. 
\Ac{aict} performs better than \ac{ict}, indicating that the introduction of a scale adaptation module can further reduce spatial redundancies and alleviate coding artifacts, especially at low bitrate resulting in higher compression efficiency.


\section{Conclusion}
\label{conclusion}
In this paper, we have proposed an up-and-coming neural codec \ac{aict}, achieving compelling \ac{rd} performance while significantly reducing the latency, which is potentially helpful to conduct, with further optimizations, high-quality real-time visual data compression. 
We inherited the advantages of self-attention units from Transformers to effectively approximate both the mean and standard deviation for entropy modeling and combine global and local texture to capture correlations among spatially neighboring components for non-linear transform coding, achieving -4.65\% BD-rate reduction over the VTM-18.0, by averaging over the benchmark datasets.
Furthermore, we presented a lightweight scale adaptation module to enhance compression ability, especially at low bitrates, reaching on average -5.11\% BD-rate reduction over the VTM-18.0.

\newpage
\bibliographystyle{IEEEbib}
\bibliography{egbib}

\end{document}